\pdfoutput=1

\documentclass[8pt]{article}

\usepackage[margin=2cm]{geometry}

\usepackage{authblk}

\usepackage{acl2023}

\usepackage{times}
\usepackage{latexsym}

\usepackage[T1]{fontenc}

\usepackage[utf8]{inputenc}

\usepackage{microtype}

\usepackage{inconsolata}

\usepackage{soul,color}
\soulregister\cite7
\soulregister\ref7
\soulregister\pageref7

\usepackage{hyperref}

\usepackage{algorithm}
\usepackage{algpseudocode}

\algdef{SE}[SUBALG]{Indent}{EndIndent}{}{\algorithmicend\ }%
\algtext*{Indent}
\algtext*{EndIndent}

\usepackage{graphicx}
\usepackage{capt-of}
\usepackage{stfloats}

\usepackage{tabularx}
\usepackage{booktabs}
\usepackage{multirow}
\usepackage[flushleft]{threeparttable}

\title{\line(1,0){500} \\[10pt] Graph-Based Retriever Captures the Long Tail of Biomedical Knowledge \\ \line(1,0){500}}

\author[]{\bf Julien Delile\thanks{\; Corresponding author: delile.julien@bcg.com}}
\author[]{\bf Srayanta Mukherjee}
\author[]{\bf Anton Van Pamel}
\author[]{\bf Leonid Zhukov}
\affil[]{Boston Consulting Group, AI Institute}

\begin{document}

\maketitle

\begin{abstract}
  Large language models (LLMs) are transforming the way information is retrieved with vast amounts of knowledge being summarized and presented via natural language conversations. Yet, LLMs are prone to highlight the most frequently seen pieces of information from the training set and to neglect the rare ones. In the field of biomedical research, latest discoveries are key to academic and industrial actors and are obscured by the abundance of an ever-increasing literature corpus (the information overload problem). Surfacing new associations between biomedical entities, e.g., drugs, genes, diseases, with LLMs becomes a challenge of capturing the long-tail knowledge of the biomedical scientific production. To overcome this challenge, Retrieval Augmented Generation (RAG) has been proposed to alleviate some of the shortcomings of LLMs by augmenting the prompts with context retrieved from external datasets. RAG methods typically select the context via maximum similarity search over text embeddings. In this study, we show that RAG methods leave out a significant proportion of relevant information due to clusters of over-represented concepts in the biomedical literature. We introduce a novel information-retrieval method that leverages a knowledge graph to downsample these clusters and mitigate the information overload problem. Its retrieval performance is about twice better than embedding similarity alternatives on both precision and recall. Finally, we demonstrate that both embedding similarity and knowledge graph retrieval methods can be advantageously combined into a hybrid model that outperforms both, enabling potential improvements to biomedical question-answering models. 
\end{abstract}

\section{Introduction}\label{introduction}

The field of biomedical research is expanding rapidly, leading to an accelerated pace of discoveries and an overwhelming surge in associated literature. Keeping track of the evolving landscape of this field is increasingly challenging for individuals. The diversity of the disciplines composing biomedical research, including molecular biology, pharmacology, clinical medicine, and epidemiology, demands an understanding of specialized terminologies and techniques that surpasses the capacity of a single individual. Amongst these disciplines, biochemistry stands out by generating a staggering number of unique entities, underlined by the indexing of over 42 million unique genes by the National Center for Biotechnology Information. These factors underscore the indispensable requirement for technologically advanced tools capable of filtering, summarizing, and elucidating this vast body of knowledge, thereby enhancing our understanding of biomedical research.

Recent years have seen breakthroughs in Large Language Models (LLMs), which have been instrumental in developing diverse applications such as question answering (QA), text summarization, language translation, and creative text generation. Amongst the methods used for producing user answers, two have been competing: (1) inclusion of information during the model training or fine-tuning phase, stored in the model weights, and (2) integrating the LLM with an external knowledge source and leveraging the model's reasoning capabilities for querying and synthesizing knowledge into comprehensible content, an approach known as Retrieval-Augmentation Generation (RAG) \cite{Lewis2020}. In the context of text summarization, we note a significant shift from the former method, generally associated with pre-trained encoder-decoder models, towards the latter \cite{Retkowski2023}, \cite{Zhang2023}, which involves large instruction-tuned autoregressive language models performing zero-shot prompting with functional tools such as LangChain \cite{Chase2022} or LlamaIndex \cite{Liu2022}. Zero-shot summarization with instruction-tuned models provides improved truthfulness, reduced hallucinations, and constant updates from an external data source, circumventing the expensive process of LLM re-training. Although standard evaluation frameworks for these tools are not yet accessible, studies suggest that the quality of LLM-generated summaries matches human-written summaries \cite{Zhang2023}. Of particular interest in the realm of summarization methodologies is Query-Based Text Summarization (QS), a unique approach that is rapidly evolving with the novel advancements in LLMs. Unlike traditional summarization, QS tailors the summary to a user-specified question \cite{Yu2022}\cite{Yang2023}. As for general text summarization, predominantly pre-trained models have been deployed for QS, as reviewed in \cite{Yu2022}, pointing towards the exploration of zero-shot approaches as a promising path for future research.

RAG systems are designed to function in two sequential stages: the extraction of targeted and pertinent text from a large magnitude of curated content, and the generation, or synthesis, of a suitable response. In this context, QA and QS are both part of a continuum of LLM tasks aiming at providing succinct and informed responses. Yet pivotal differences between these tasks lies within the abundance of information retrieved from the text corpus and the complexity involved in the synthesis phase. For QA tasks, the response may be found directly within the text corpus, initiating the retrieval of narrow and specific text \textquotesingle chunks\textquotesingle{} that are rephrased by the LLM to formulate the answer. Consequently, the precision of retrieval is paramount to deliver accurate and valid information. In contrast, QS tasks often require a broader pull of information that covers a wider spectrum of the query\textquotesingle s nuances. Enhancements to the LLM\textquotesingle s reasoning capability, coupled with an increased context length (up to 32K for GPT4 and 100k for Claude2), enables the ability to respond to more comprehensive queries across an extensive corpus such as Pubmed.

However, retrieving larger amount of information into the synthesizer context presents a distinct challenge that need to be addressed. Current LLM models are struggling to fully exploit all the retrieved content, e.g., multi-document question-answering performance is degraded as the context grows longer \cite{Liu2023}. While new attention-based neural architectures designed to deal with very large context window may prevent performance drops \cite{Yu2023}, an efficient selection of the most relevant information has become critical, not least for the sake of latency, cost, and energy consumption. This challenge is particularly prevalent in the context of biomedical literature which contains extensively redundant pieces of information. This text corpus presents an \emph{information overload} problem, where rare and recent yet important information that can be dominated by over-represented older concepts.

In this study, leaving aside the generative side of RAG, we introduce a novel knowledge-graph-based retrieval approach that enables access to the long tail of biomedical knowledge. We demonstrate that RAG retrieval approaches, typically based on text chunk embedding similarity, leave out a significant proportion of relevant information because of the data imbalance in a queried text corpus such as Pubmed. Some over-represented topics can preclude the RAG synthesizer to access more recent discoveries by monopolizing the list of most similar text chunks. We propose to perform a rebalancing of the retrieved text chunks by under-sampling these larger clusters of information, and to do so by structuring the text corpus with a knowledge graph of biomedical entities (genes, diseases and diseases). In addition, our method also provides control mechanisms to prioritize the retrieval of recent and impactful discoveries. Finally, we built a hybrid approach combining the strengths of LLM embedding semantic relationships and structured knowledge graph and show that it outperforms both embedding similarity and knowledge graph based methods for biomedical information retrieval.

\begin{figure*}[b]  
  \includegraphics[width=\linewidth]{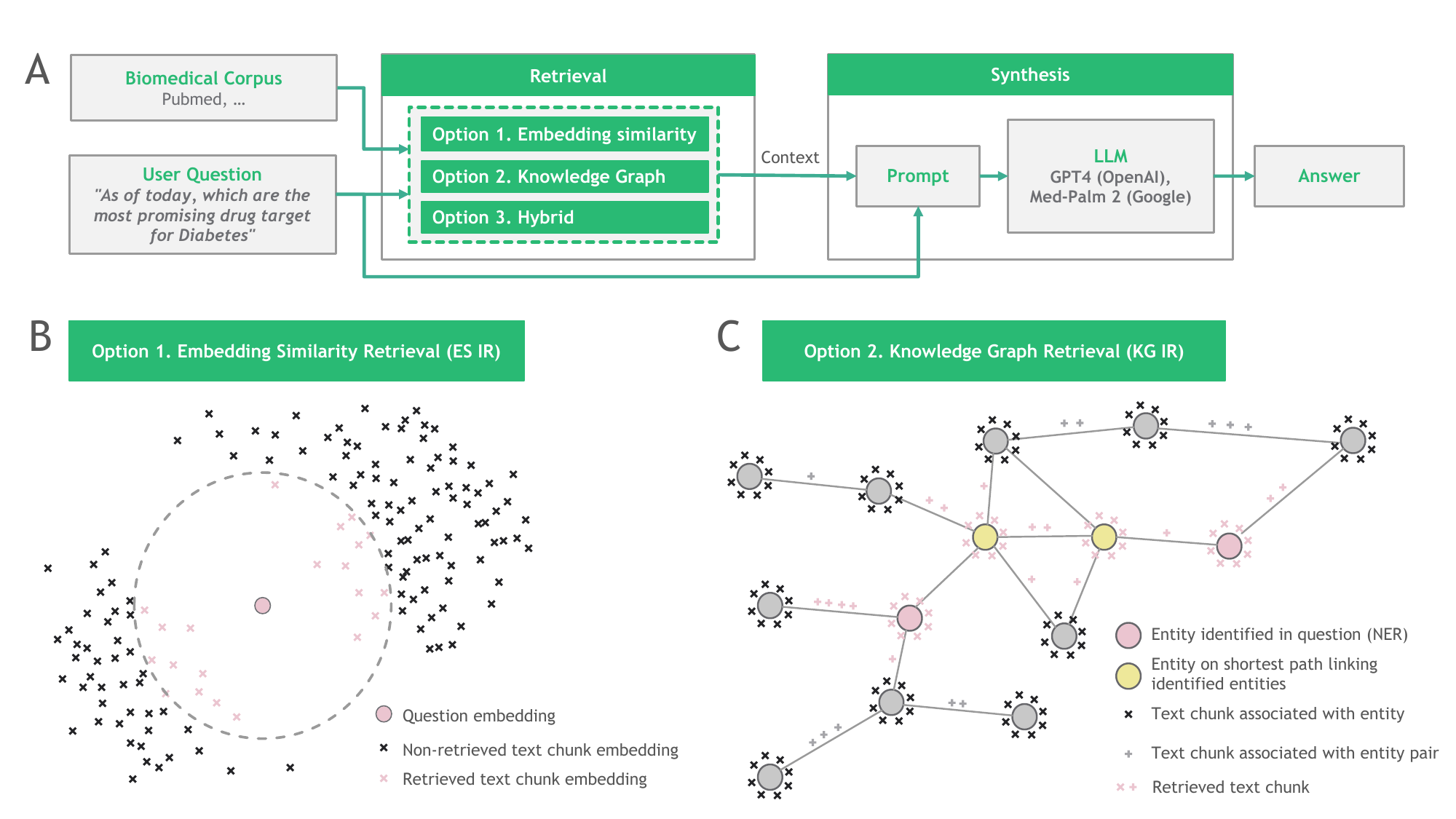}
  \captionof{figure}{Knowledge Graphs enable an alternative retrieval-augmentation mechanism. A. Retrieval-augmentation enables question-answering from knowledge unseen during LLM training by inserting additional text chunks into the prompt (Retrieval) and summarising the augmented prompt following to the user question. B. A commonly used retrieval mechanism involves selecting text chunks from the corpus which are the most similar to the user question in embedding space (e.g. the top-20 closest neighbours using cosine similarity and OpenAI's \emph{text-embedding-ada-002} embedding engine). C. Proposed knowledge graph-based IR approach maps text chunks to the graph nodes and edges and use entity recognition in the question to retrieve the texts along the shortest path linking the identified entities.}
  \label{Fig:mechanism}
  \end{figure*}

\section{Related work}\label{related-work}

Knowledge Graphs (KGs) are structured knowledge models linking real-world or abstract concepts, or \emph{entities}, with relationships in the form (\emph{head entity}, \emph{relation}, \emph{tail entity}). KGs have been proposed to complement LLMs in various settings such as hallucination limitations \cite{Ji2022}\cite{Feng2023} and LLM interpretability \cite{Lin2019}. As reviewed in \cite{Pan2023}, KGs have also been combined with LLMs to enhance pre-training and inference \cite{Zhang2019}\cite{Yasunaga2021}, and reciprocally LLMs have been leveraged to completement KG tasks such as entity embedding \cite{Zhang2020}, link prediction \cite{Yao2019}\cite{Xie2022} and KG construction \cite{Bosselut2019}\cite{Han2023}.

To solve multiple-choice question-answering tasks, a group of models have combined question-answer text encodings with Graph Neural Network embeddings (GNN) of KG paths linking entities present in the question and candidate answers. The joint embedding of each candidate question-answer pair is attributed a plausibility score via a multilayer perceptron. KagNet \cite{Lin2019} combines graph convolutional networks \cite{Kipf2016} and LSTMs \cite{Hochreiter1997} to solve multiple-choice question answering. It feeds KG multi-hop paths linking pairs of question entities and candidate answer entities into an attention-based GCN to attribute a plausibility score to each pair. Similarly, MHGRN \cite{Feng2020}, QA-GNN \cite{Yasunaga2021}, JointLK \cite{Sun2021} and GreaseLM \cite{Zhang2022} combines GNN embeddings of the KG question-answer entity subgraphs with a LLM embedding of the question produced by an encoder-only transformer LLM such as RoBERTA \cite{Liu2019}. These hybrid graph-text encoding models are all trained on multiple-choice QA datasets and lack capabilities to generate natural language responses to queries.

Pre-transformer-era model KGLM is an autoregressive language model leveraging LSTMs to define a probability distribution for the next token \cite{Logan2019}. Each predicted token is taken among the entities of a knowledge graph by mapping the "prompt" sequence to the most likely parent entity and relation. KGLM is also able to learn a local knowledge graph made of new entities if unseen in its training set. This approach is reminiscent of the knowledge graph index implemented in the LlamaIndex package \cite{Liu2022}. A KG index is built by extracting (\emph{subject}, \emph{predicate}, \emph{object}) triplets from a text corpus using an autoregressive LLM (defaulting to OpenAI completion endpoint). For each user question, keywords are then extracted using the same LLM and used to retrieve triplets in the k-hop neighbourhood of the keywords. The retrieved triplets are then added to the synthesizer context for answer generation.

To the best of our knowledge, this study is the first to highlight the information overload problem in the context of text chunk embedding similarity retrieval and to propose a graph-distance retrieval approach to mitigate its effect.

\section{Methods}\label{methods}

A typical information retrieval (IR) workflow is composed of two sequential steps: the retrieval step and the synthesis step. The former aims at retrieving text chunks that share a relationship with the user question and the latter combine the retrieved context and the user question in its prompt to produce an answer (Fig. 1A). In this section, we detail two alternative approaches to perform the retrieval step:

\begin{itemize}
\item
  \begin{quote}
  IR using a similarity function between dense embeddings of the user question and the text corpus (Fig. 1B).
  \end{quote}
\item
  \begin{quote}
  IR using a novel Knowledge-Graph approach relying on entity recognition and relationship extraction performed with model trained and fine-tuned for biomedical literature (Fig 1C, Fig 2).
  \end{quote}
\end{itemize}

\subsection{IR with text embedding similarity}\label{ir-with-text-embedding-similarity}

We use text Embedding Similarity Information Retrieval (ES IR) as a baseline approach for this study.

\subsubsection{Embedding indexing}\label{embedding-indexing}

We built an embedding index from a subset of the \textasciitilde35M articles available on Pubmed (NCBI FTP site\footnote{\href{https://ftp.ncbi.nlm.nih.gov/pubmed/}{https://ftp.ncbi.nlm.nih.gov/pubmed/}}). Only the article having an abstract were retained.

For each experiment, a different subset was used and is specified in the associated experiment section (typically, about 100k documents were indexed per experiment).

In all cases, each selected article's title and abstract were split into individual sentences using \emph{en\_core\_sci\_md}, a sentence tokenizer trained on large biomedical dataset \cite{Neumann2019}.

Embeddings for each sentence were obtained using OpenAI's second generation embedding model \emph{text-embedding-ada-002}. Each embedding is encoded into a 1536-dimension vector. While little is known about OpenAI's embedding model specifications, \emph{text-embedding-ada-002 }ranks among the top-8 retrieval text embedding models on scientific facts benchmark (e.g. SciFact benchmark on MTEB\footnote{\href{https://huggingface.co/spaces/mteb/leaderboard}{https://huggingface.co/spaces/mteb/leaderboard}}), offers a larger input size (8191 tokens) and showed clear semantic pattern on the datasets analysed in this study (Fig. 4A-B). In very rare occasions (\textless.01\%), a sentence would be longer than the embedding model input token limit. Those sentences were split to fit the limit.

\subsubsection{Retrieval}\label{retrieval}

While complex approaches have been developed, e.g. MaxSim \cite{Khattab2020}, most ES IR strategies rely on either cosine/inner-product or Euclidean similarity function to retrieve the text chunks whose embeddings are the most similar to the user question embedding vector \cite{Wang2022}. In this study, we use cosine similarity to rank the embedded text chunks for each query. Various retrieval rank thresholds were used in this study and are specified in the experiment section.

\subsection{IR with knowledge graph support}\label{ir-with-knowledge-graph-support}

The rationale behind using a knowledge graph for IR is that traditional text embedding similarity approaches are limited by the imbalance of pieces of information in a large corpus of text such as Pubmed. Some topics have been documented in greater length than others. While topics are often over-represented because of their importance for a field information (e.g. the EGFR drug target is referenced in more than 84k+ articles related to cancer), they can hide other relevant information by their sheer number when semantic similarity is used for retrieval.

Rebalancing the retrievable text chunks can be done by undersampling the larger clusters of information. The problem then becomes how to define these clusters. While these clusters could be defined by the text chunk distributions in semantic space, a domain knowledge like biomedical literature has produced vast amounts of ontologies covering all types of entities that can be used to organise the information. We propose to leverage this existing knowledge to undersample clusters of information built around three types of biomedical entities (genes, diseases and chemical compounds).

\subsubsection{Building the biomedical Knowledge Graph}\label{building-the-biomedical-knowledge-graph}

To build the knowledge graph that aim at structuring and rebalancing the information stored in the biomedical corpus, we used two frameworks sequentially:

\begin{itemize}
\item
  \begin{quote}
  The KAZU framework was used to extract entities: gene, diseases and chemical compounds\cite{Yoon2022}. It includes the NER model TinyBERN2, a computationally efficient version of the BERN2 NER model \cite{Sung2022}. KAZU then performs entity normalization, a critical step that link the multiple variations of a single entity to a reference vocabulary provided by the following ontologies: Ensembl (genes), MONDO (diseases) and ChEMBL (chemical compounds). Finally, it also disambiguates and/or merges overlapping candidate entities in input text chunks.
  \end{quote}
\item
  \begin{quote}
  The PubmedBERT model \cite{Gu2020} fine-tuned on the BioRED dataset to perform relationship extraction (RE) from scientific abstracts\cite{Luo2022}. PubmedBERT is a transformer-based encoder LLM with BERT architecture that was pre-trained on Pubmed abstracts and PMC full -text articles. PubmedBERT was then fine-tuned on BioRED, a recently published dataset containing \textasciitilde6.5k curated relationships between biomedical entities over 600 abstracts. Pairs of disease, gene or chemical compound entities annotated by KAZU are linked in a knowledge graph if the RE model predicts a relationship between them. Six types of relationships are predicted: association, positive correlation, negative correlation, cotreatment, comparison and bind.
  \end{quote}
\end{itemize}

\subsubsection{Knowledge Graph Indexing}\label{knowledge-graph-indexing}

We mapped all the text chunks produced for the embedding index onto the nodes and edges of the knowledge graph (Fig. 2). The following rules were applied to perform the mapping:

\begin{itemize}
\item
  \begin{quote}
  Only text chunks with at least one annotation are mapped
  \end{quote}
\item
  \begin{quote}
  Text chunks with a single annotated entity are associated with the node of that entity
  \end{quote}
\item
  \begin{quote}
  Text chunks with two annotated entities are associated with the corresponding edge iff the pair has been labelled by the RE model
  \end{quote}
\item
  \begin{quote}
  Text chunks with two annotated entities are associated with both entity nodes iff the pair has not been labelled by the RE model
  \end{quote}
\item
  \begin{quote}
  Previous two rules are applied to all combinatorial entity pairs in text chunks with 3+ entities
  \end{quote}
\end{itemize}

\begin{figure*}[t]  
  \includegraphics[width=\linewidth]{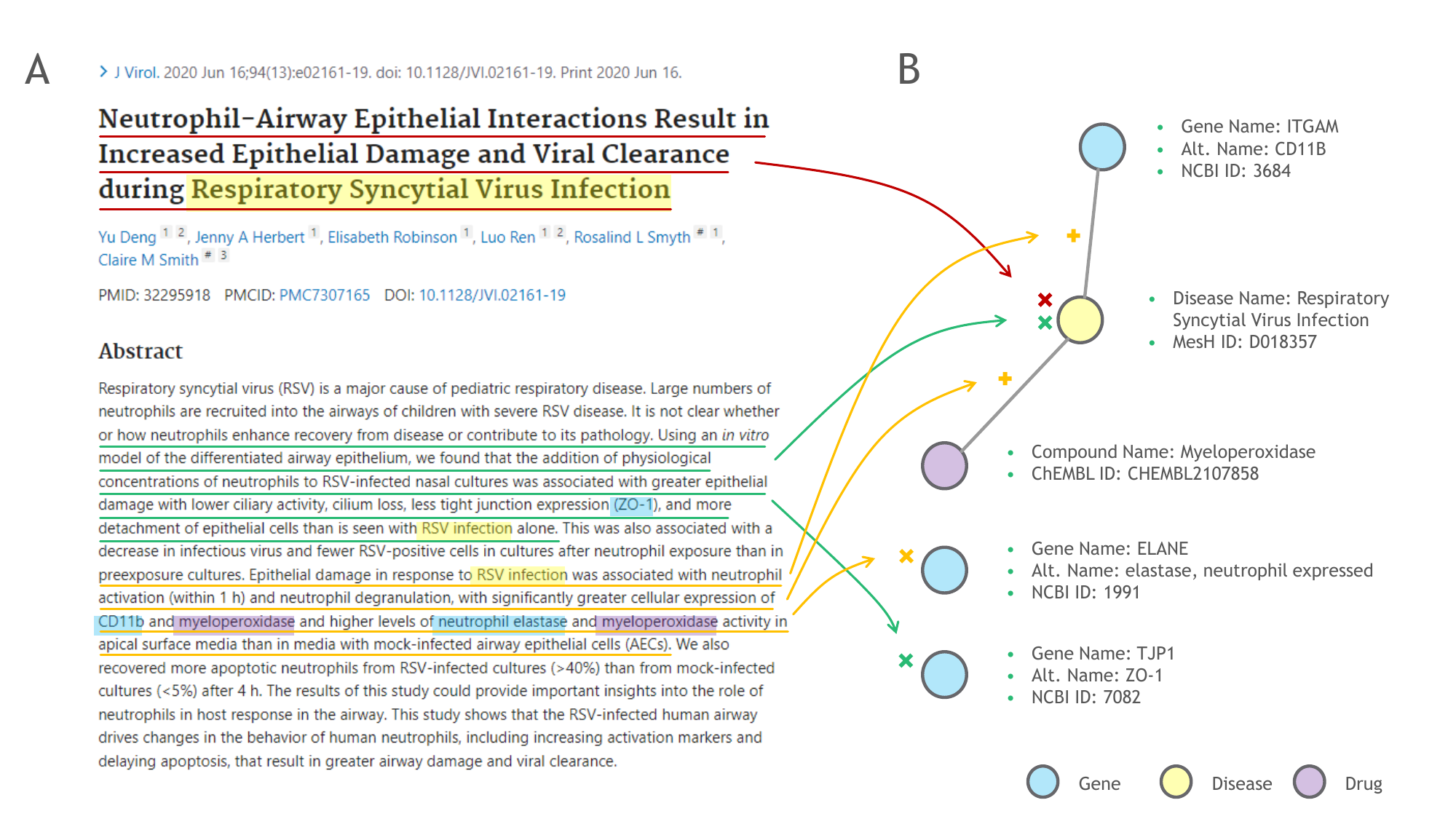}
  \captionof{figure}{Annotations from scientific articles structure the biomedical knowledge graph used for retrieval. A. A scientific article is annotated by extracting entities (NER) and the type of relationships linking entities (RE). B. Each sentence containing 1+ entity is mapped onto the knowledge graph, either on a link related two entities if it contains two entities that are linked semantically ($+$ sign) or to its individual entities otherwise ($\times$ sign).}
  \label{Fig:annotations}
\end{figure*}

\subsubsection{Retrieval}\label{retrieval-1}

Once text chunks have been mapped to the knowledge graph, we can exploit graph distances to retrieve the chunks that are the most relevant to the user question. The first step is to identify which entity(ies) are the starting point of the graph-based retrieval. We leverage the KAZU pipeline to identify which entities are present in the user question. We then build the shortest path relating these entities in the graph and retrieve text chunks mapped to the shortest path entities and their neighbour edges. This approach allows the synthesizer LLM to produce non-trivial answers. For example, if a question asks to explain the relationship between two entities whose interaction has not been directly documented in the literature, text chunks from intermediate entities are presented and allows to build a undirect explanation that may potentially lead to new discoveries. Yet, this approach by itself is still incomplete as it lacks a text chunks prioritization metric. Indeed, hundreds of thousands of text chunks may be mapped along the shortest path in the graph.

To prioritize the most relevant text chunks and enable a rebalancing of the data that gives a fair chance to each concept mapped along the shortest path, we introduce a scoring metric that factor in both the recency and the impact of a text chunk. We measure the impact of a text chunk as the total number of citations that the associated document received. Because recent articles have less citations but are more likely to contain new discoveries, we solve the trade-off between these two objectives by using the pareto front of the recency/impact space. The algorithm is detailed in Algorithm \ref{alg:prio}.

\begin{algorithm*}
\caption{Retrieval Prioritization Ranking}\label{alg:prio}
\begin{algorithmic}[1]
\Require K -- number of retrieved text chunks
\Ensure K text chunks along with a ranking score
\State Set ranking score variable S to 1
\State Until the total number of retrieved text chunks K is reached
\Indent
    \State For each node and all their edges along the graph shortest path linking the question entities
    \Indent
        \State Select text chunks mapped to current node/edge
        \State Select documents associated with these text chunks
        \State Identify documents on the Pareto front for both recency and impact objectives
        \State Collect text chunks associated with Pareto optimal documents
    \EndIndent
    \State Set ranking score of collected text chunks to S
    \State Increment S by 1
    \State Remove collected text chunks from pool of retrievable text chunks
\EndIndent
\end{algorithmic}
\end{algorithm*}

The ranking algorithm combined with the graph-distance approach enables to satisfy the main criterion we identified as key for improving upon embedding similarity retrieval, i.e. data rebalancing giving an equal chance to each entity related to a question, while also surfacing latest significant discoveries.

Optional adjustable exclusion thresholds for minimum year of publication and minimum number of publications have been implemented in order to favour higher quality source of information. These filters are not used in this study as we should apply them simultaneously to the embedding index for a fair comparison.

\section{Experiments}\label{experiments}

\subsection{Comparing knowledge graph and embedding similarity retrieval performance}\label{comparing-knowledge-graph-and-embedding-similarity-retrieval-performance}

Here, we aim at comparing the retrieval performance of both embedding similarity (ES IR) and knowledge graph (KG IR) retrieval strategies. We purposefully use an open question that requires to explore a wide range of documents: \emph{``What are the known drug targets for treating \textless disease\textgreater?''} and compare the retrieved information of both approaches with curated annotations produced by biomedical experts. We repeat the experiment over 8 diseases selected to cover the different therapeutic areas included in the dataset: asthma, pulmonary arterial hypertension, heart failure, hypertension, Parkinson's disease, Alzheimer's disease, liver cirrhosis, inflammatory bowel disease.

\begin{figure*}[t]  
  \includegraphics[width=\linewidth, trim = 0cm 10cm 6.5cm 0cm, clip]{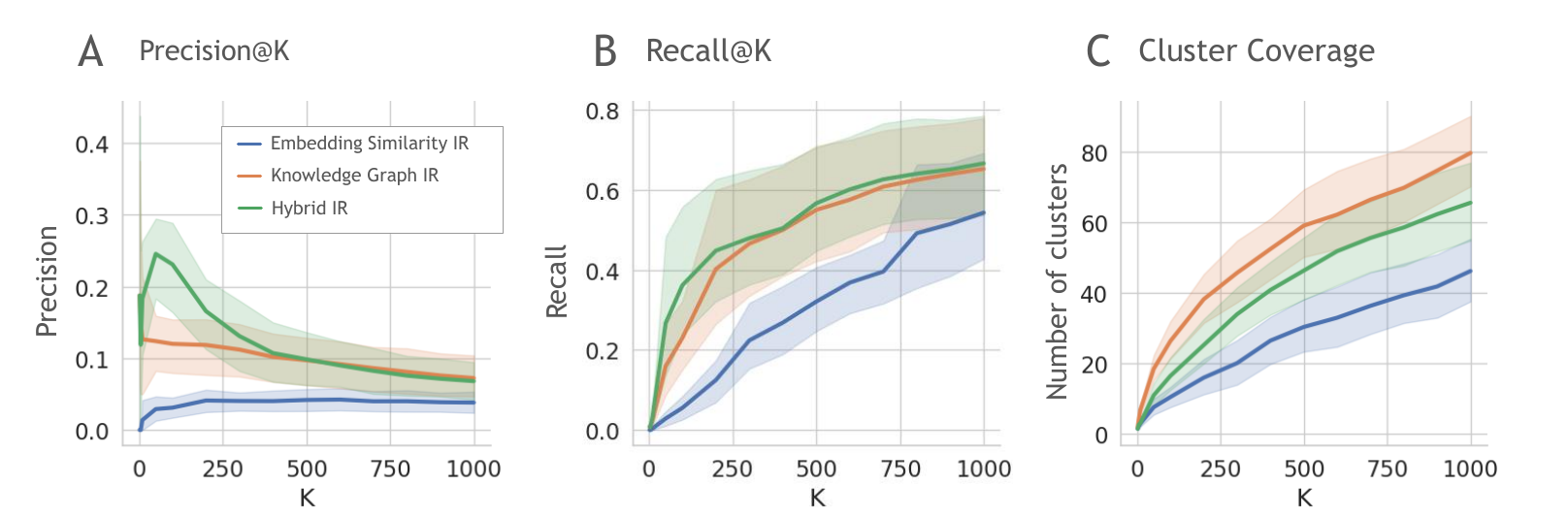}
  \captionof{figure}{Retrieval performance comparison between embedding similarity IR (blue), knowledge graph IR (orange) and hybrid method (green). The metrics compare each method's retrieval performance for the same task: retrieving biomedical text chunks which are relevant to the question ``What are the known drug targets for treating \textless disease\textgreater?'' over 8 diseases: asthma, pulmonary arterial hypertension, heart failure, hypertension, Parkinson's disease, Alzheimer's disease, liver cirrhosis, inflammatory bowel disease. Solid lines indicate the metric averages and transparent ribbon 95\% confidence intervals. A-B. Recall@K and Precision@K. C. Number of clusters containing at least one retrieved text chunk, among the 200 clusters defined in 1536-dimension embedding space.}
  \label{Fig:performance}
\end{figure*}

\textbf{Text corpus.} To present a wide landscape of topics mixing relevant and irrelevant information, we produced both embedding and knowledge graph indexes using Pubmed articles belonging to five therapeutic areas (nervous system, respiratory tract diseases, digestive system diseases, cardiovascular diseases, mental disorders). About 1\% of the articles were sampled randomly to produce a corpus of 86166 articles and 731238 sentences. Both embeddings and knowledge graph indexes were generated as described in sections 2.1 and 2.2. The embedding index is ordered by cosine similarity between the question embedding and the text chunk embeddings. The knowledge graph index is built and ordered by using the 372999 text chunks mapped to the graph entities or entity edges. For each question, the 1-hop neighbourhood of the question's disease entity, i.e. the entity node and the edges towards the neighbour entities, is available for retrieval as only one entity is present in the question. A breakdown of the 1-hop neighbourhood for each disease in provided in Table 1.

\begin{table*}[pb]
\centering
\caption{Gold-standard and local knowledge graph metrics for each disease explored in the retrieval method comparison}
\label{table:graph-metrics}
\begin{threeparttable}
\begin{tabularx}{\linewidth}{l *{7}{c}}
\toprule
\multirow{2}{*}{\textbf{Disease}} &
\multirow{2}{*}{\textbf{Therapeutic Area}} & 
\multirow{2}{*}{\textbf{Entity MeSH ID}} & 
\multicolumn{3}{c}{\textbf{Curated Dataset}} & 
\multicolumn{2}{c}{\textbf{KG 1-hop neighborhood}} \\
& & & \# Articles & \# Chunks & \# Genes & \# Articles & \# Chunks \\
\midrule
\textbf{Asthma} & Respiratory Tract & D001249 & 41 & 442 & 27 & 1278 & 3670 \\
\textbf{PAH\tnote{a}} & Respiratory Tract & D000081029 & 2 & 26 & 2 & 498 & 1358 \\
\textbf{Heart Failure} & Cardiovascular & D006333 & 37 & 400 & 21 & 1878 & 4351 \\
\textbf{Hypertension} & Cardiovascular & D006973 & 79 & 845 & 18 & 2681 & 5031 \\
\textbf{Parkinson} & Nervous System & D010300 & 61 & 553 & 23 & 962 & 3022 \\
\textbf{Alzheimer} & Nervous System & D000544 & 115 & 1083 & 53 & 1503 & 4547 \\
\textbf{Liver Cirrhosis} & Digestive System & D008103 & 49 & 582 & 19 & 863 & 1798 \\
\textbf{IBD\tnote{b}} & Digestive System & D015212 & 57 & 642 & 22 & 561 & 1586 \\
\bottomrule
\end{tabularx}
\begin{tablenotes}
\item[a] PAH: Pulmonary Arterial Hypertension.
\item[b] IBD: Inflammatory Bowel Disease.
\end{tablenotes}
\end{threeparttable}
\end{table*}

\textbf{Gold-standard dataset.} For each question, we compare the retrieved documents with a list of documents annotated by subject experts and containing both an annotation for the disease and for at least one gene in its full text. Two annotation sources were leveraged. First, the National Library of Medicine (NLM) provides a manually curated list of Medical Subject Headings (MeSH) terms for each article indexed in Pubmed/MEDLINE \cite{Aronson2000}. We selected disease-related items from their tree number (e.g. starting with \emph{C08.127.108} for Asthma and its children diseases in MeSH). Secondly, we mapped the potential drug targets for Asthma using GeneRIF annotations\footnote{\href{https://www.ncbi.nlm.nih.gov/gene/about-generif}{https://www.ncbi.nlm.nih.gov/gene/about-generif}}. GeneRIF allows scientists to upload functional annotation of genes that are then reviewed by life science experts. On average, \textasciitilde23 genes were identified in the subset of articles mapped to the question disease, leading to a gold-standard dataset of \textasciitilde55 documents split into \textasciitilde571 embeddings (Table \ref{table:graph-metrics}).

We note that both types of curated annotations are not covering exhaustively all articles in the corpus, particularly for genes as scientists submit GeneRIF annotations on a voluntary basis. Yet, while this leads to an underestimation of the performance of both retrieval mechanisms which may retrieve relevant but unannotated documents, we assume that this does not favour one approach against the other hence allows fair comparison.

\textbf{Results.} To assess the performance of both retrieval mechanisms, we adopt two metrics widely used for information retrieval and recommender systems: precision@K and recall@K. Because annotations are available at the document level and not at the text chunk level, we calculate both metrics by counting the number of retrieved documents rather than retrieved text chunks. A document is considered retrieved if at least one of its text chunks is retrieved.

Overall, KG IR strongly outperforms ES IR on both metrics. As expected from the incomplete nature of the gold-standard dataset, the precision is low in both models. As the retrieval window increases from K=0 to K=1000, ES IR reaches a peak value of \textasciitilde5\% at K=250 while KG IR progressively decreases from \textasciitilde12\% to \textasciitilde8\% (Fig. 3B). Less affected by missing, yet relevant, documents in the gold-standard dataset, recall in KG IR\,shows a large gap of performance over ES IR (Fig. 3C). Nearly 43\% of the gold-standard documents are retrieved by KG IR\,when K=250 text chunks are queried. In contrast, ES IR has only 17\% of the gold-standard documents for the same retrieval volume. We observe ES IR\,recall keeps on increasing and eventually exceeds KG IR recall for a very large set of retrieved documents (K \textgreater{} 1000). This is because the KG IR approach only retrieves documents that have been mapped by NER to the knowledge graph while ES IR has access to the whole embedded corpus.

\subsection{Knowledge graph retrieval accesses the long tail of biomedical knowledge}\label{knowledge-graph-retrieval-accesses-the-long-tail-of-biomedical-knowledge}

To explain the gap in performance between both approaches, we hypothesized that KG IR is able to access the long-tail knowledge of the corpus that embedding similarity approaches are missing. We investigated this hypothesis by comparing the distribution of retrieved information for both methods with the gold-standard dataset in embedding space.

\textbf{Landscape projection}. To visualise the distributions of retrieved information over the experiment text corpus, we used a non-linear dimension reduction technique, Uniform Manifold Approximation and Projection (UMAP), on all \textasciitilde731k 1536-dimensional embedding vectors. UMAP learns a projection transformation that aims at maintaining the samples' local neighbourhood in low dimensional space (here 2D). It was selected because of its computational efficiency on large dataset and ability to be applied on new text embeddings. Indeed, once trained and applied to the text corpus, we also transformed the embedding of one of the questions from the previous experiment (\emph{What are the known drug targets for treating Asthma?}), allowing us to compare the question location within the text corpus.

\textbf{Retrieved text localisation}. To highlight the location of the retrieved text chunks in embedding space, we query the top-200 text chunks for both ES IR and KG IR methods. We apply Gaussian kernels (bandwidth factor set to 0.25) onto the UMAP coordinates of the retrieved text chunks in order to estimate the probability density function of retrieval for both methods. We then visualise the higher density regions in UMAP space by drawing filled contours for p \textgreater{} 0.02 (Fig. 4D). To assess the spread of the retrieved chunks in embedding space, we performed a k-means clustering of the \textasciitilde731k chunk embeddings (k=200) and counted the number of clusters containing at least one chunk for all retrieval parameters (Fig. 3C).

\textbf{Results}. Overall, the projected embedding landscape presents many high-density regions corresponding to over-represented concepts in the corpus (Fig. 4A-E). This is explained by the broad range of biomedical concepts stored in the Pubmed corpus subset. To assess the relevance and complexity of the landscape, we mapped the regions linked to the five covered disease areas and the various types of entities stored in the text chunks. The landscapes reveal an clear pattern of localised disease areas in embedding space (ordered from top-left to bottom-right in Fig. 4B), as well as an orthogonal pattern for the types of entities expressed in each text chunk (concentric region centred around genes top-right corner, followed by chemical compounds/drugs and diseases at the periphery Fig. 4A).

To assess whether ES IR can link the question with a diverse range of concepts, we overlayed the landscape with the cosine similarity between the question embedding and the text corpus embeddings. This reveals that most of the most similar text chunks are localized in the region surrounding the question embedding (blue cross in Fig. 4C), but also that different parts of the landscape are semantically linked to the question (black arrows in Fig. 4C). This indicates that the poor performance of ES IR is not due to its inability to build non-trivial semantic relationships but rather to access the longtail knowledge.

We hypothesized that the cause is rather the lack of data balancing that makes ES IR retrieve text chunks predominantly from the closest high-density region. This hypothesis is supported by two observations. First, comparing the region of ES IR retrieved text chunks (blue region in Fig. 4D) and the distribution of gold-standard embedding (grey dots), we observe that the ES IR retrieval region is densely localized in the vicinity of the question embedding. In contrast, KG IR retrieval region is multipolar and covers a wider range of curated documents (Fig. 4D). A more granular comparison of the retrieved articles' text chunks in the landscape shows that KG IR also captures other smaller clusters of curated documents that were not part of the dense retrieval regions (Fig. 4E). Second, we quantified the spread of retrieval in embedding space by counting the number of k-means clusters that each retrieved chunk sets belong to (Fig. 3C). For the same retrieval volume, ES IR reaches less than half the number of clusters compared to KG IR.

Both granular and regional observations lead us to the conclusion that, in contrast to ES IR, the data balancing mechanism of KG IR allows it to go beyond the immediate surrounding of the question neighborhood to retrieve relevant information, hence is facilitating the capture of the longtail knowledge of biomedical information.

\begin{figure*}[bp]  
  \centering
  \includegraphics[width=\linewidth]{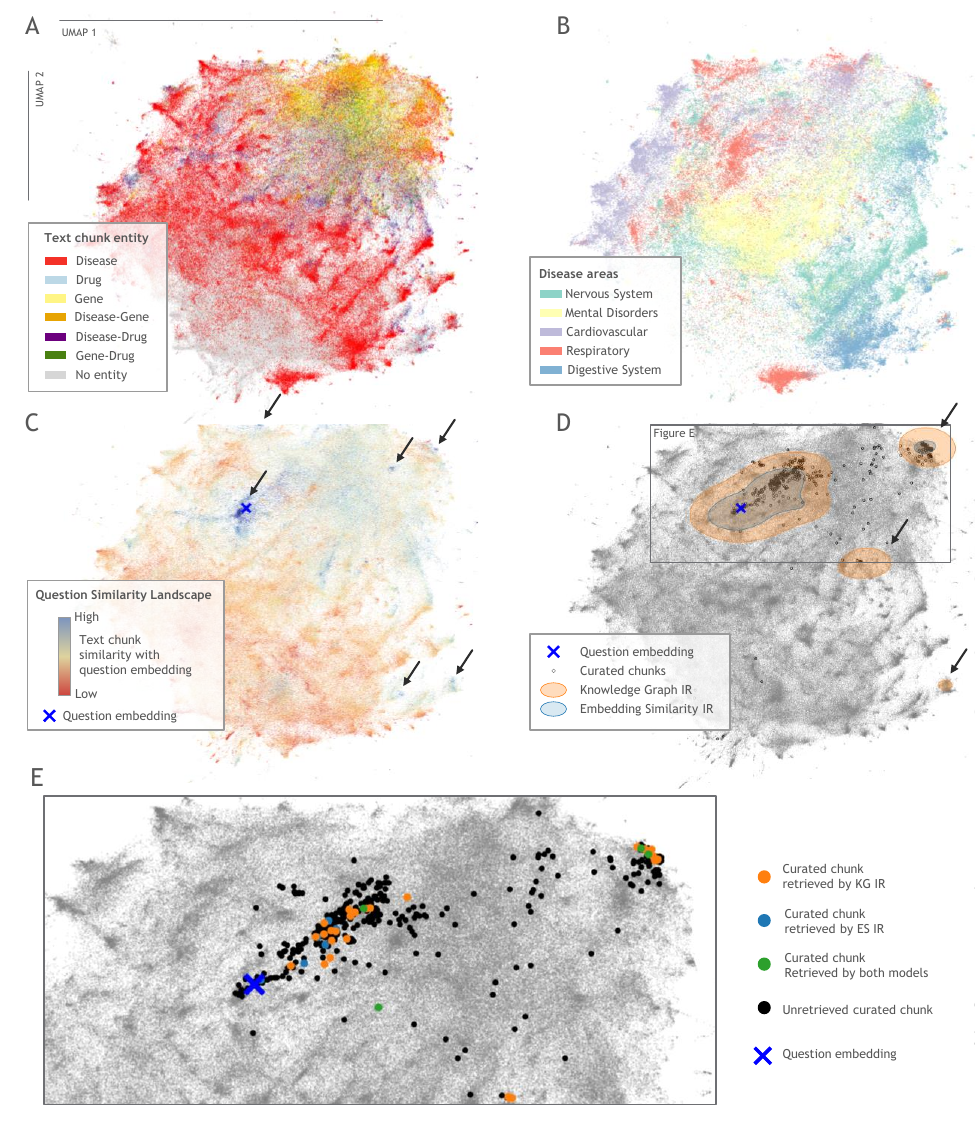}
  \captionof{figure}{Characterization of differences between the ES IR and KG IR methods over the text embedding landscape. Each plot represents the \textasciitilde731k 1536-dimensional text chunk embeddings in two dimensions via UMAP transformation. A. The biomedical entity landscape illustrates the entity(ies) present in each text chunk. Text chunks containing a pair of genes (resp. drugs) are represented with same colour as text chunks containing a single gene (resp. drug). B. Disease area landscape. Only text chunks containing 1+ disease entity are represented. C. Question similarity landscape, colours indicate the cosine similarity of the text chunk embedding with the question embedding (``What are the known drug targets for treating Asthma?'', also in D-E). Arrows indicate remote spots of text chunks that are most similar to the question embedding. D. High-density retrieval regions indicates the parts of the landscape where both methods are retrieving most of their chunks for K=200. Arrow indicates a secondary cluster of high-density retrieval for KG IR. Black dots represent the 355 text chunks that are part of the gold-standard dataset. E. Granular comparison of the retrieved documents for K=200. All dots represent the 442 curated text chunks.}
  \label{Fig:landscape}
\end{figure*}

\subsection{Semantic and graph-based retrieval approaches are highly complementary}\label{semantic-and-graph-based-retrieval-approaches-are-highly-complementary}

Here we attempt to combine the strengths of both methods in order to maximize the retrieval performance.

\textbf{Hybrid ranking}. As both ES IR and KG IR offers different strengths to perform retrieval, and both produce quantitative ranking scores, we evaluate a hybrid approach (Hybrid IR) that build upon those methods. For a single retrieval task, both embedding similarity and KG retrieval prioritization ranking scores of all available text chunks are first rescaled between 0 and 1\,independently for each method. Then each text chunk's pair of rescaled scores is averaged to produce a new ranking score. This new score is then used to retrieve the top-k text chunks for a user question. We note that only the text chunks mapped to the knowledge graph are available for Hybrid IR as the KG prioritization ranking score is required.

\textbf{Datasets}. We use the same indexed dataset, curated gold-standard dataset and question as for the first experiment described in 3.1.

\textbf{Results. }Adopting the same performance evaluation metrics as in 3.1., we observe that Hybrid IR strongly outperforms both ES IR and KG IR for smaller volume of retrieved information (K\textless100) and moderately when the retrieval window increases beyond K=250 (Fig.3A-B). Both recall and precision are about twice higher for Hybrid IR compared to KG IR for K=50. This indicates that each base retrieval method provides a complementary mechanism: data rebalancing from KG IR is not sufficient to identify the most relevant pieces of information and benefits from adding the semantic filter provided by embedding similarity.

\section{Discussion}\label{discussion}

Considering the rapid rise of LLMs in various tasks such as question answering and text summarization \cite{Yu2022}\cite{Yang2023}, the problem of RAG for highly specialized domains like biomedical information is a challenging yet critical field of research. To maximize the reasoning ability of LLMs and to aid contextual synthesis, larger and more diverse information retrieval plays a pre-eminent role to maintain a balanced and unbiased selection of retrieved information. Hence, extracting the long tail of biomedical information is of utmost importance. Here, we study the strengths of a knowledge-graph based information retrieval system, compare it against the more standard vector embedding similarity-based retrieval system. Our findings strongly suggest that the KG based system has significantly better performance overall but is also complementary to the embedding similarity approach and therefore a hybrid system of retrieval outperforms both methods individually.

Our findings further show that the tendency of ES IR to oversample the immediate neighbourhood of the question embedding due to its lack of data balancing. On the other hand, the presence of a mechanism in KG to balance the data enables the search to extend beyond the immediate neighbourhood and thereby identify a more diverse set of relevant documents. This fundamental difference between the retrieval mechanism of these two methods, also therefore spawns mutual complementarity leading to a hybrid approach being superior to both individual ones.

It has been shown in earlier research, that a knowledge fusion into LLMs using a KG based RAG, improves generation performance versus baseline by helping the text synthesis to be more factual, diverse, and yet specific \cite{Logan2019} \cite{Guu2020}. At this point we do not test the effect of accessing the biomedical long-tail information on answer generation, but it can be reasonably hypothesized that performance of content generation should also see an improvement as observed in the open-domain. However, one important point to consider and evaluate during content generation will be effect of decreasing signal/noise ratio when accessing the long tail. In our study, the precision of KG IR, while quite low is still better than ES IR, hence careful controls can possibly be placed into the RAG system to maintain performance.

Another possible mechanism to improve the performance of RAG would be to leverage LLM ``chain of thoughts (COT)''. Considering that several neighbouring entities tend to be irrelevant to the question, COT can used to de-clutter the neighbourhood and direct entity mapping onto the KG and thereby retrieve additional contextual information. Particularly, for more complex questions, containing multiple entities that may require multi-hop retrieval (or sub-graph identification), COT directed RAG may provide additional. While not a subject of this study, it provides one possible future direction for research. Note, while we restricted our study to the biomedical information domain, similar KG based RAG approaches are an active area of research with open-domain retrieval as well \cite{Zhou2018}.

Another general area of research in the joint KG+LLM space, is KG-to-text generation, that consistently and accurately describe the information embedded in the KG. While applicability of KG-to-text in more real-world natural language generation scenarios are many \cite{Ribeiro2020} \cite{Ke2021}, it is difficult and expensive to collect graph-text parallel corpus, which in turn results in low training fidelity leading to poor quality generation. In our study, we constructed a weakly supervised KG starting from journal articles in free text formal which can therefore be extended to create a large, though domain specific, graph-text parallel corpus. Thus, one possible future direction of research would be reverse our current application of RAG to direct generation of text when provided with relevant graph entities. The accurate generation from the KG would be of immense help in applications such as automated authoring of scientific articles, regulatory submissions etc.

\bibliographystyle{jae}

\bibliography{biblio}

\begin{thebibliography}{39}
\expandafter\ifx\csname natexlab\endcsname\relax\def\natexlab#1{#1}\fi
\expandafter\ifx\csname url\endcsname\relax
  \def\url#1{\texttt{#1}}\fi
\expandafter\ifx\csname urlprefix\endcsname\relax\def\urlprefix{URL }\fi

\bibitem[{Aronson et~al.(2000)Aronson, Bodenreider, Chang, Humphrey, Mork,
  Nelson, Rindflesch and Wilbur}]{Aronson2000}
Aronson AR, Bodenreider O, Chang HF, Humphrey SM, Mork JG, Nelson SJ,
  Rindflesch TC, Wilbur WJ. 2000.
\newblock The nlm indexing initiative.
\newblock \emph{Proceedings of the AMIA Symposium} : 17ISSN 1531605X.
\newline\urlprefix\url{/pmc/articles/PMC2243970/?report=abstract
  https://www.ncbi.nlm.nih.gov/pmc/articles/PMC2243970/}

\bibitem[{Bosselut et~al.(2019)Bosselut, Rashkin, Sap, Malaviya, Celikyilmaz
  and Choi}]{Bosselut2019}
Bosselut A, Rashkin H, Sap M, Malaviya C, Celikyilmaz A, Choi Y. 2019.
\newblock Comet: Commonsense transformers for automatic knowledge graph
  construction.
\newblock \emph{ACL 2019 - 57th Annual Meeting of the Association for
  Computational Linguistics, Proceedings of the Conference} : 4762--4779.
\newline\urlprefix\url{https://arxiv.org/abs/1906.05317v2}

\bibitem[{Chase(2022)}]{Chase2022}
Chase H. 2022.
\newblock Langchain.
\newline\urlprefix\url{https://github.com/hwchase17/langchain}

\bibitem[{Feng et~al.(2023)Feng, Balachandran, Bai and Tsvetkov}]{Feng2023}
Feng S, Balachandran V, Bai Y, Tsvetkov Y. 2023.
\newblock Factkb: Generalizable factuality evaluation using language models
  enhanced with factual knowledge .
\newline\urlprefix\url{https://arxiv.org/abs/2305.08281v1}

\bibitem[{Feng et~al.(2020)Feng, Chen, Lin, Wang, Yan and Ren}]{Feng2020}
Feng Y, Chen X, Lin BY, Wang P, Yan J, Ren X. 2020.
\newblock Scalable multi-hop relational reasoning for knowledge-aware question
  answering.
\newblock \emph{EMNLP 2020 - 2020 Conference on Empirical Methods in Natural
  Language Processing, Proceedings of the Conference} : 1295--1309.
\newline\urlprefix\url{https://arxiv.org/abs/2005.00646v2}

\bibitem[{Gu et~al.(2020)Gu, Tinn, Cheng, Lucas, Usuyama, Liu, Naumann, Gao and
  Poon}]{Gu2020}
Gu Y, Tinn R, Cheng H, Lucas M, Usuyama N, Liu X, Naumann T, Gao J, Poon H.
  2020.
\newblock Domain-specific language model pretraining for biomedical natural
  language processing.
\newblock \emph{ACM Transactions on Computing for Healthcare} \textbf{3}: 24.
\newline\urlprefix\url{http://arxiv.org/abs/2007.15779
  http://dx.doi.org/10.1145/3458754}

\bibitem[{Guu et~al.(2020)Guu, Lee, Tung, Pasupat and Chang}]{Guu2020}
Guu K, Lee K, Tung Z, Pasupat P, Chang MW. 2020.
\newblock Realm: Retrieval-augmented language model pre-training.
\newblock \emph{37th International Conference on Machine Learning, ICML 2020}
  \textbf{PartF168147-6}: 3887--3896.
\newline\urlprefix\url{https://arxiv.org/abs/2002.08909v1}

\bibitem[{Han et~al.(2023)Han, Collier, Buntine and Shareghi}]{Han2023}
Han J, Collier N, Buntine W, Shareghi E. 2023.
\newblock Pive: Prompting with iterative verification improving graph-based
  generative capability of llms .
\newline\urlprefix\url{https://arxiv.org/abs/2305.12392v1}

\bibitem[{Hochreiter and Schmidhuber(1997)}]{Hochreiter1997}
Hochreiter S, Schmidhuber J. 1997.
\newblock Long short-term memory.
\newblock \emph{Neural Computation} \textbf{9}: 1735--1780.
\newblock ISSN 0899-7667.
\newline\urlprefix\url{https://dx.doi.org/10.1162/neco.1997.9.8.1735}

\bibitem[{Ji et~al.(2022)Ji, Liu, Lee, Yu, Wilie, Zeng and Fung}]{Ji2022}
Ji Z, Liu Z, Lee N, Yu T, Wilie B, Zeng M, Fung P. 2022.
\newblock Rho ($\rho$): Reducing hallucination in open-domain dialogues with
  knowledge grounding .
\newline\urlprefix\url{https://arxiv.org/abs/2212.01588v2}

\bibitem[{Ke et~al.(2021)Ke, Ji, Ran, Cui, Wang, Song, Zhu and Huang}]{Ke2021}
Ke P, Ji H, Ran Y, Cui X, Wang L, Song L, Zhu X, Huang M. 2021.
\newblock Jointgt: Graph-text joint representation learning for text generation
  from knowledge graphs.
\newblock \emph{Findings of the Association for Computational Linguistics:
  ACL-IJCNLP 2021} : 2526--2538.
\newline\urlprefix\url{https://arxiv.org/abs/2106.10502v1}

\bibitem[{Khattab and Zaharia(2020)}]{Khattab2020}
Khattab O, Zaharia M. 2020.
\newblock Colbert: Efficient and effective passage search via contextualized
  late interaction over bert.
\newblock \emph{SIGIR 2020 - Proceedings of the 43rd International ACM SIGIR
  Conference on Research and Development in Information Retrieval} : 39--48.
\newline\urlprefix\url{https://arxiv.org/abs/2004.12832v2}

\bibitem[{Kipf and Welling(2016)}]{Kipf2016}
Kipf TN, Welling M. 2016.
\newblock Semi-supervised classification with graph convolutional networks.
\newblock \emph{5th International Conference on Learning Representations, ICLR
  2017 - Conference Track Proceedings} .
\newline\urlprefix\url{https://arxiv.org/abs/1609.02907v4}

\bibitem[{Lewis et~al.(2020)Lewis, Perez, Piktus, Petroni, Karpukhin, Goyal,
  Küttler, Lewis, Yih, Rocktäschel, Riedel and Kiela}]{Lewis2020}
Lewis P, Perez E, Piktus A, Petroni F, Karpukhin V, Goyal N, Küttler H, Lewis
  M, Yih WT, Rocktäschel T, Riedel S, Kiela D. 2020.
\newblock Retrieval-augmented generation for knowledge-intensive nlp tasks.
\newblock \emph{Advances in Neural Information Processing Systems}
  \textbf{2020-December}.
\newblock ISSN 10495258.
\newline\urlprefix\url{https://arxiv.org/abs/2005.11401v4}

\bibitem[{Lin et~al.(2019)Lin, Chen, Chen and Ren}]{Lin2019}
Lin BY, Chen X, Chen J, Ren X. 2019.
\newblock Kagnet: Knowledge-aware graph networks for commonsense reasoning.
\newblock \emph{EMNLP-IJCNLP 2019 - 2019 Conference on Empirical Methods in
  Natural Language Processing and 9th International Joint Conference on Natural
  Language Processing, Proceedings of the Conference} : 2829--2839.
\newline\urlprefix\url{https://arxiv.org/abs/1909.02151v1}

\bibitem[{Liu(2022)}]{Liu2022}
Liu J. 2022.
\newblock Llamaindex.
\newline\urlprefix\url{https://github.com/jerryjliu/gpt_index}

\bibitem[{Liu et~al.(2023)Liu, Lin, Hewitt, Paranjape, Bevilacqua, Petroni and
  Liang}]{Liu2023}
Liu NF, Lin K, Hewitt J, Paranjape A, Bevilacqua M, Petroni F, Liang P. 2023.
\newblock Lost in the middle: How language models use long contexts .
\newline\urlprefix\url{https://arxiv.org/abs/2307.03172v2}

\bibitem[{Liu et~al.(2019)Liu, Ott, Goyal, Du, Joshi, Chen, Levy, Lewis,
  Zettlemoyer, Stoyanov and Allen}]{Liu2019}
Liu Y, Ott M, Goyal N, Du J, Joshi M, Chen D, Levy O, Lewis M, Zettlemoyer L,
  Stoyanov V, Allen PG. 2019.
\newblock Roberta: A robustly optimized bert pretraining approach .
\newline\urlprefix\url{https://arxiv.org/abs/1907.11692v1}

\bibitem[{Logan et~al.(2019)Logan, Liu, Peters, Gardner and Singh}]{Logan2019}
Logan RL, Liu NF, Peters ME, Gardner M, Singh S. 2019.
\newblock Barack's wife hillary: Using knowledge-graphs for fact-aware language
  modeling.
\newblock \emph{ACL 2019 - 57th Annual Meeting of the Association for
  Computational Linguistics, Proceedings of the Conference} : 5962--5971.
\newline\urlprefix\url{https://arxiv.org/abs/1906.07241v2}

\bibitem[{Luo et~al.(2022)Luo, Lai, Wei, Arighi and Lu}]{Luo2022}
Luo L, Lai PT, Wei CH, Arighi CN, Lu Z. 2022.
\newblock Biored: a rich biomedical relation extraction dataset.
\newblock \emph{Briefings in Bioinformatics} \textbf{23}.
\newblock ISSN 1467-5463.
\newline\urlprefix\url{https://academic.oup.com/bib/article/doi/10.1093/bib/bbac282/6645993}

\bibitem[{Neumann et~al.(2019)Neumann, King, Beltagy and Ammar}]{Neumann2019}
Neumann M, King D, Beltagy I, Ammar W. 2019.
\newblock Scispacy: Fast and robust models for biomedical natural language
  processing.
\newblock \emph{BioNLP 2019 - SIGBioMed Workshop on Biomedical Natural Language
  Processing, Proceedings of the 18th BioNLP Workshop and Shared Task} :
  319--327.
\newline\urlprefix\url{http://arxiv.org/abs/1902.07669
  http://dx.doi.org/10.18653/v1/W19-5034}

\bibitem[{Pan et~al.(2023)Pan, Member, Luo, Wang, Chen, Wang and Wu}]{Pan2023}
Pan S, Member S, Luo L, Wang Y, Chen C, Wang J, Wu X. 2023.
\newblock Unifying large language models and knowledge graphs: A roadmap .
\newline\urlprefix\url{https://arxiv.org/abs/2306.08302v1}

\bibitem[{Retkowski(2023)}]{Retkowski2023}
Retkowski F. 2023.
\newblock The current state of summarization .
\newline\urlprefix\url{https://arxiv.org/abs/2305.04853v2}

\bibitem[{Ribeiro et~al.(2020)Ribeiro, Schmitt, Schütze and
  Gurevych}]{Ribeiro2020}
Ribeiro LF, Schmitt M, Schütze H, Gurevych I. 2020.
\newblock Investigating pretrained language models for graph-to-text
  generation.
\newblock \emph{NLP for Conversational AI, NLP4ConvAI 2021 - Proceedings of the
  3rd Workshop} : 211--227.
\newline\urlprefix\url{https://arxiv.org/abs/2007.08426v3}

\bibitem[{Sun et~al.(2021)Sun, Shi, Qi and Zhang}]{Sun2021}
Sun Y, Shi Q, Qi L, Zhang Y. 2021.
\newblock Jointlk: Joint reasoning with language models and knowledge graphs
  for commonsense question answering.
\newblock \emph{NAACL 2022 - 2022 Conference of the North American Chapter of
  the Association for Computational Linguistics: Human Language Technologies,
  Proceedings of the Conference} : 5049--5060.
\newline\urlprefix\url{https://arxiv.org/abs/2112.02732v2}

\bibitem[{Sung et~al.(2022)Sung, Jeong, Choi, Kim, Lee and Kang}]{Sung2022}
Sung M, Jeong M, Choi Y, Kim D, Lee J, Kang J. 2022.
\newblock Bern2: an advanced neural biomedical named entity recognition and
  normalization tool.
\newblock \emph{Bioinformatics} \textbf{38}: 4837--4839.
\newblock ISSN 1367-4803.
\newline\urlprefix\url{https://academic.oup.com/bioinformatics/article/38/20/4837/6687126}

\bibitem[{Wang(2022)}]{Wang2022}
Wang Y. 2022.
\newblock A survey on efficient processing of similarity queries over neural
  embeddings .
\newline\urlprefix\url{https://arxiv.org/abs/2204.07922v1}

\bibitem[{Xie et~al.(2022)Xie, Zhang, Li, Deng, Chen, Xiong, Chen and
  Chen}]{Xie2022}
Xie X, Zhang N, Li Z, Deng S, Chen H, Xiong F, Chen M, Chen H. 2022.
\newblock From discrimination to generation: Knowledge graph completion with
  generative transformer.
\newblock \emph{WWW 2022 - Companion Proceedings of the Web Conference 2022} :
  162--165.
\newline\urlprefix\url{http://arxiv.org/abs/2202.02113
  http://dx.doi.org/10.1145/3487553.3524238}

\bibitem[{Yang et~al.(2023)Yang, Li, Zhang, Chen and Cheng}]{Yang2023}
Yang X, Li Y, Zhang X, Chen H, Cheng W. 2023.
\newblock Exploring the limits of chatgpt for query or aspect-based text
  summarization .
\newline\urlprefix\url{http://arxiv.org/abs/2302.08081}

\bibitem[{Yao et~al.(2019)Yao, Mao and Luo}]{Yao2019}
Yao L, Mao C, Luo Y. 2019.
\newblock Kg-bert: Bert for knowledge graph completion .
\newline\urlprefix\url{https://arxiv.org/abs/1909.03193v2}

\bibitem[{Yasunaga et~al.(2021)Yasunaga, Ren, Bosselut, Liang and
  Leskovec}]{Yasunaga2021}
Yasunaga M, Ren H, Bosselut A, Liang P, Leskovec J. 2021.
\newblock Qa-gnn: Reasoning with language models and knowledge graphs for
  question answering.
\newblock \emph{NAACL-HLT 2021 - 2021 Conference of the North American Chapter
  of the Association for Computational Linguistics: Human Language
  Technologies, Proceedings of the Conference} : 535--546.
\newline\urlprefix\url{https://arxiv.org/abs/2104.06378v5}

\bibitem[{Yoon et~al.(2022)Yoon, Jackson, Ford, Poroshin and Kang}]{Yoon2022}
Yoon W, Jackson R, Ford E, Poroshin V, Kang J. 2022.
\newblock Biomedical ner for the enterprise with distillated bern2 and the kazu
  framework .
\newline\urlprefix\url{https://arxiv.org/abs/2212.00223v1}

\bibitem[{Yu(2022)}]{Yu2022}
Yu H. 2022.
\newblock Survey of query-based text summarization .
\newline\urlprefix\url{https://arxiv.org/abs/2211.11548v1}

\bibitem[{Yu et~al.(2023)Yu, Simig, Flaherty, Aghajanyan, Zettlemoyer and
  Lewis}]{Yu2023}
Yu L, Simig D, Flaherty C, Aghajanyan A, Zettlemoyer L, Lewis M. 2023.
\newblock Megabyte: Predicting million-byte sequences with multiscale
  transformers .
\newline\urlprefix\url{https://arxiv.org/abs/2305.07185v2}

\bibitem[{Zhang et~al.(2023)Zhang, Ladhak, Durmus, Liang, McKeown and
  Hashimoto}]{Zhang2023}
Zhang T, Ladhak F, Durmus E, Liang P, McKeown K, Hashimoto TB. 2023.
\newblock Benchmarking large language models for news summarization .
\newline\urlprefix\url{https://arxiv.org/abs/2301.13848v1}

\bibitem[{Zhang et~al.(2022)Zhang, Bosselut, Yasunaga, Ren, Liang, Manning and
  Leskovec}]{Zhang2022}
Zhang X, Bosselut A, Yasunaga M, Ren H, Liang P, Manning CD, Leskovec J. 2022.
\newblock Greaselm: Graph reasoning enhanced language models for question
  answering.
\newblock \emph{ICLR 2022 - 10th International Conference on Learning
  Representations} .
\newline\urlprefix\url{https://arxiv.org/abs/2201.08860v1}

\bibitem[{Zhang et~al.(2019)Zhang, Han, Liu, Jiang, Sun and Liu}]{Zhang2019}
Zhang Z, Han X, Liu Z, Jiang X, Sun M, Liu Q. 2019.
\newblock Ernie: Enhanced language representation with informative entities.
\newblock \emph{ACL 2019 - 57th Annual Meeting of the Association for
  Computational Linguistics, Proceedings of the Conference} : 1441--1451.
\newline\urlprefix\url{https://arxiv.org/abs/1905.07129v3}

\bibitem[{Zhang et~al.(2020)Zhang, Liu, Zhang, Su, Sun and He}]{Zhang2020}
Zhang Z, Liu X, Zhang Y, Su Q, Sun X, He B. 2020.
\newblock Pretrain-kge: Learning knowledge representation from pretrained
  language models.
\newblock \emph{Findings of the Association for Computational Linguistics
  Findings of ACL: EMNLP 2020} : 259--266.
\newline\urlprefix\url{https://aclanthology.org/2020.findings-emnlp.25}

\bibitem[{Zhou et~al.(2018)Zhou, Young, Huang, Zhao, Xu and Zhu}]{Zhou2018}
Zhou H, Young T, Huang M, Zhao H, Xu J, Zhu X. 2018.
\newblock Commonsense knowledge aware conversation generation with graph
  attention.
\newblock \emph{IJCAI International Joint Conference on Artificial
  Intelligence} \textbf{2018-July}: 4623--4629.
\newblock ISSN 10450823.

\end{thebibliography}

\end{document}